\pdfoutput=1
\documentclass[11pt]{article}

\usepackage[margin=1in]{geometry}
\usepackage{amsmath,amssymb,mathtools}
\usepackage{graphicx}
\usepackage{booktabs}
\usepackage{caption} 
\usepackage{float}
\usepackage[hidelinks]{hyperref}
\usepackage{amsmath,amssymb,mathtools}
\usepackage{tabularx,array,booktabs}
\usepackage{graphicx}
\usepackage{booktabs}
\usepackage{caption} 
\usepackage{float}
\usepackage{placeins} 
\usepackage{tabularx,makecell}

\usepackage[final]{microtype}
\PassOptionsToPackage{hyphens}{url} 
\usepackage{doi}

\usepackage{algorithm}
\usepackage{algorithmic}

\title{A Hybrid Enumeration Framework for Optimal Counterfactual Generation in Post-Acute COVID-19 Heart Failure}
\author{
\begin{tabular}{c}
Jingya Cheng, M.B.$^{1*\dagger}$;\; Alaleh Azhir, MD, MSc$^{1,2*}$;\;Jiazi Tian, M.Sc.$^{1}$;\;  \\ Hossein Estiri, PhD$^{1}$
\end{tabular}
}
\date{
\small
$^{1}$Clinical Augmented Intelligence Group, Massachusetts General Hospital, Boston, MA, USA \\
$^{2}$Department of Medicine, Brigham and Women's Hospital, Boston, MA, USA \\
*Equal contribution (co–first authors)  \quad $\dagger$Corresponding: \texttt{jcheng30@mgh.harvard.edu} \\[3pt]
399 Revolution Drive, Suite 790, Somerville, MA, 02145, USA
}

\begin{document}

\maketitle

\begin{abstract}
Counterfactual inference provides a mathematical framework for reasoning about hypothetical outcomes under alternative interventions, bridging causal reasoning and predictive modeling. We present a counterfactual inference framework for individualized risk estimation and intervention analysis, illustrated through a clinical application to post-acute sequelae of COVID-19 (PASC) among patients with pre-existing heart failure (HF). Using longitudinal diagnosis, laboratory, and medication data from a large health-system cohort, we integrate regularized predictive modeling with counterfactual search to identify actionable pathways to PASC-related HF hospital admissions. The framework combines exact enumeration with optimization-based methods, including the Nearest Instance Counterfactual Explanations (NICE) and Multi-Objective Counterfactuals (MOC) algorithms, to efficiently explore high-dimensional intervention spaces. Applied to more than 2700 individuals with confirmed SARS-CoV-2 infection and prior HF, the model achieved strong discriminative performance (AUROC = 0.88, 95\% CI: 0.84–0.91) and generated interpretable, patient-specific counterfactuals that quantify how modifying comorbidity patterns or treatment factors could alter predicted outcomes. This work demonstrates how counterfactual reasoning can be formalized as an optimization problem over predictive functions, offering a rigorous, interpretable, and computationally efficient approach to personalized inference in complex biomedical systems.
\end{abstract}

\section{Introduction}
Patients with pre-existing heart failure (HF) who survive COVID-19 remain at elevated risk for post-acute sequelae of COVID-19 (PASC), including recurrent or newly emergent HF-related hospitalizations \cite{soriano2021pasc,xie2022cardio,wang2022eclin,salah2022hf,mohamed2022nrc}. Despite growing recognition of PASC as a heterogeneous syndrome affecting multiple organ systems, the specific and modifiable determinants of vulnerability in patients with underlying HF remain poorly understood \cite{soriano2021pasc,mohamed2022nrc}. Identifying actionable insights is essential for designing targeted prevention strategies and improving long-term outcomes in this high-risk population \cite{xie2022cardio,wang2022eclin}.

Addressing this challenge requires analytical approaches that move beyond population-level associations to capture individualized mechanisms of disease progression. Traditional statistical and machine learning models typically estimate risk factors at the cohort level, overlooking the heterogeneity in diagnostic histories, treatment responses, and comorbidity interactions across patients. However, each individual’s clinical trajectory reflects a unique combination of exposures and physiological responses, demanding models capable of reasoning about personalized outcomes and intervention effects \cite{hernan2020}.

Counterfactual inference provides an interpretable framework for addressing this gap. By estimating “what-if” scenarios, it bridges predictive modeling and causal reasoning \cite{hernan2020} to understand how outcomes might change under hypothetical interventions. Counterfactual approaches enable researchers to evaluate not only which features are predictive of an outcome, but also how modifying those features could alter an individual’s predicted risk \cite{shalit2017,nguyen2020}. This perspective transforms conventional risk modeling into a tool for personalized decision support and mechanistic insight, offering a forward-looking view of disease trajectories rather than retrospective correlation \cite{wachter2018,brughmans2021nice,dandl2020moc}.

Here, we present a data-driven framework that integrates regularized predictive modeling with counterfactual inference to identify actionable contributors to PASC-related HF admissions. Using temporal diagnostic, laboratory, and medication data from a large precision research cohort,\cite{azhir2024p2rc} our approach estimates individualized risk trajectories and quantifies the potential impact of modifying key clinical features. Through this integration, counterfactual reasoning serves not only as an interpretability mechanism, but also as a computational framework for simulating personalized interventions. The resulting model provides patient-specific, interpretable insights into modifiable risk factors, illustrating how counterfactual inference can advance both individualized prediction and causal understanding in complex clinical settings.

\section{Method}
We used longitudinal electronic health record (EHR) from the validated Precision Post-Acute Sequelae of COVID-19 Research Cohort (P2RC) \cite{azhir2024p2rc} from Mass General Brigham (MGB). P2RC comprises clinical data from over 85,000 patients with at least a confirmed COVID-19 infection across eight hospitals and affiliated community health centers across Massachusetts.

Eligible participants were those with a documented COVID-19 diagnosis between March 6, 2020, and June 7, 2022, and with subsequent follow-up extending through June 7, 2023. The analytic design followed a retrospective EHR-based cohort approach to evaluate clinical trajectories and to identify modifiable determinants of post-infection outcomes. Patients meeting criteria for Post-Acute Sequelae of COVID-19 (PASC) within one year of infection were labeled as PASC cases, whereas those without sequelae during the same period served as non-PASC comparators.

Within the MGB P2RC, all patients with pre-existing heart failure who subsequently developed 
post-acute sequelae of COVID-19 (PASC-HF) were identified and matched to 
COVID-19–positive patients with pre-existing heart failure controls without PASC at a ratio of 1:6. 
Matching was performed by age group, sex, Elixhauser Comorbidity Index (ECI), 
and the time of SARS-CoV-2 infection to control to account for demographic and temporal confounding.

Model development employed the \textit{Temporal Learning with Dynamic Range (TLDR)}  \cite{cheng2025tldr}, a hard-attention operator that maps EHR timelines into a compact, interpretable set of temporal features by querying fixed windows and selecting salient events. By embedding temporal relevance directly into the feature space, TLDR complements our counterfactual framework by providing temporal, clinically meaningful inputs. 
We trained a regularized gradient-boosted model with five-fold cross-validation 
using diagnosis, laboratory, and medication records 
from MGB patients with prior HF and confirmed COVID-19 infection. 
The primary outcome was PASC-related HF admission, defined according to 
World Health Organization (WHO) criteria for PASC. 
The dataset was split 70/30 into training and testing subsets. 
Model discrimination was quantified using the area under the receiver operating characteristic curve (AUROC), 
and the optimal classification threshold was determined using Youden’s index.

Finally, counterfactual inference was applied to estimate the potential impact of modifiable clinical factors on predicted risk, enabling simulation of “what-if” scenarios that illustrate how changes in comorbidities or treatment variables could alter individual-level outcomes.

\subsection{Hybrid Counterfactual Generation Framework}
We propose a novel hybrid algorithm that adaptively selects between exhaustive enumeration and optimization-based search depending on the number of binary actionable features. The key innovation lies in recognizing that binary features, those indicating presence or absence of attributes, admit complete enumeration when their cardinality is small, guaranteeing discovery of optimal sparse counterfactuals.

Let $\mathcal{X} \subseteq \mathbb{R}^p$ denote the feature space with $p$ features, and let $f: \mathcal{X} \rightarrow [0,1]^C$ be a trained classifier that outputs probability distributions over $C$ classes. Given an instance of interest $\mathbf{x}_0 \in \mathcal{X}$ with predicted class $\hat{y}_0 = \arg\max_c f_c(\mathbf{x}_0)$ and a desired target class $y^* \neq \hat{y}_0$, our objective is to find counterfactual explanations $\mathbf{x}' \in \mathcal{X}$ such that:

\begin{equation}
f_{y^*}(\mathbf{x}') \in [p_{\min}, p_{\max}], \quad \text{where } 0 \leq p_{\min} < p_{\max} \leq 1
\end{equation}

subject to domain-specific constraints. We partition the feature set into fixed features $\mathcal{F} \subset \{1, \ldots, f\}$ that must remain unchanged and actionable features $\mathcal{A} = \{1, \ldots, a\} \setminus \mathcal{F}$ that may be modified.

\subsection{Binary Feature Identification and Enumeration}
Let $\mathcal{B} \subseteq \mathcal{A}$ denote the subset of binary actionable features. A feature $j$ is classified as binary if it takes exactly two distinct values, typically representing presence/absence, yes/no, or true/false conditions:

\begin{equation}
j \in \mathcal{B} \iff |\{x_{i,j} : \mathbf{x}_i \in \mathcal{D}\}| = 2
\end{equation}

where $\mathcal{D}$ is the feature dataset. Let $m = |\mathcal{B}|$ denote the count of binary actionable features.

The algorithm branches based on $m$ relative to a user-specified threshold $m_{\max}$ ($m \leq m_{\max}$). When the number of binary actionable features is sufficiently small, we perform exhaustive enumeration over all $2^m - 1$ non-trivial feature combinations. The enumeration strategy systematically explores the complete space of binary feature modifications to guarantee discovery of the optimal sparse counterfactual.

For each non-empty subset $S \subseteq \mathcal{B}$, we construct a candidate counterfactual $\mathbf{x}'_S$ by toggling the binary features indexed by $S$ while preserving all other features:

\begin{equation}
x'_{S,j} = \begin{cases}
1 - x_{0,j} & \text{if } j \in S \text{ and } j \in \mathcal{B} \\
x_{0,j} & \text{otherwise}
\end{cases}
\end{equation}

assuming binary features are encoded as $\{0, 1\}$. This generates exactly $2^m - 1$ distinct candidates (excluding the trivial no-change case where $S = \emptyset$).

Each candidate $\mathbf{x}'_S$ is evaluated in two steps. We first verify whether the candidate satisfies the target probability constraint:
\begin{equation}
p_S = f_{y^*}(\mathbf{x}'_S) \in [p_{\min}, p_{\max}]
\end{equation}

Only candidates passing this constraint are retained for further evaluation. Valid candidates are then scored using a composite function that balances sparsity and effectiveness:
\begin{equation}
\mathcal{S}(\mathbf{x}'_S) = \alpha \cdot |S| - \beta \cdot |f_{y^*}(\mathbf{x}'_S) - f_{y^*}(\mathbf{x}_0)|
\end{equation}
where $|S|$ denotes the number of binary features changed (sparsity measure), $\alpha > 0$ is the penalization weight for complexity, $\beta > 0$ is the reward weight for probability shift toward the target class, and $f_{y^*}(\mathbf{x}_0)$ is the current probability of the target class for the original instance.

The scoring function $\mathcal{S}(\cdot)$ embodies two competing objectives: minimizing $|S|$ promotes interpretability by reducing the number of feature changes, while maximizing $|f_{y^*}(\mathbf{x}'_S) - f_{y^*}(\mathbf{x}_0)|$ ensures substantial probability shift toward the target class. The relative importance of these objectives is controlled by the ratio $\beta/\alpha$.

All valid candidates are ranked by ascending score (lower is better), and the top $k$ are selected as counterfactual explanations. This enumeration approach provides a \emph{global optimality guarantee}: when $m \leq m_{\max}$, the returned counterfactuals are provably optimal with respect to $\mathcal{S}(\cdot)$ within the binary feature subspace.

The enumeration branch evaluates exactly $2^m - 1$ candidates, each requiring one prediction from $f$. To maintain computational tractability, we enforce $m \leq m_{\max} = 16$, limiting the worst-case to $2^{16} - 1 = 65{,}535$ evaluations. When $2^m - 1$ exceeds $2^{20} \approx 10^6$, the algorithm automatically falls back to optimization methods to prevent combinatorial explosion.

\subsection{Optimization-Based Fallback}
When enumeration is computationally infeasible or when actionable features are predominantly continuous or high-cardinality categorical, the algorithm employs optimization-based methods. The optimization fallback employs a hierarchical two-stage approach, attempting first NICE and then MOC if necessary.

\paragraph{NICE (Nearest Instance Counterfactual Explanations).} NICE searches for training instances of the desired class that are nearest to $\mathbf{x}_0$ \cite{brughmans2021nice}. To respect fixed feature constraints $\mathcal{F}$, we restrict the candidate pool to training instances that match $\mathbf{x}_0$ on all fixed features. NICE computes distances using the Gower metric to handle mixed-type feature spaces and returns the $k$ nearest neighbors whose predicted probabilities satisfy Equation~(1). The method is particularly effective when the training data contains naturally occurring instances similar to $\mathbf{x}_0$ that already belong to the target class.

\paragraph{MOC (Multi-Objective Counterfactuals).} If NICE fails to produce valid counterfactuals, either due to insufficient restricted training samples or constraint infeasibility, we employ MOC \cite{dandl2020moc}, an algorithm that explores synthetic variants. MOC uses a genetic algorithm framework to simultaneously optimize multiple objectives: proximity to $\mathbf{x}_0$, sparsity (number of changed features), and target class probability. The algorithm maintains a population of candidate solutions and evolves it over multiple generations using crossover and mutation operators adapted for mixed-type features. Fixed features in $\mathcal{F}$ are explicitly protected from modification during the evolutionary process. The final counterfactuals are selected from the Pareto-optimal front, representing the best trade-offs among the competing objectives.

Both NICE and MOC outputs undergo post-hoc validation to ensure all constraints are satisfied: fixed features must remain at their original values within numerical tolerance ($\epsilon = 10^{-8}$), and only actionable features in $\mathcal{A}$ may be modified.

\subsection{Algorithm Summary}
\begin{algorithm}[H]
\caption{Hybrid Counterfactual Generation for Classification}
\label{alg:hybrid_cf}
\begin{algorithmic}[1]
\REQUIRE Predictor $f$, instance $\mathbf{x}_0$, target class $y^*$, probability range $[p_{\min}, p_{\max}]$
\REQUIRE Fixed features $\mathcal{F}$, max counterfactuals $k$, enum threshold $m_{\max}$
\REQUIRE Scoring weights $\alpha$ (penalty), $\beta$ (reward)
\ENSURE Set of counterfactuals $\mathcal{C} = \{\mathbf{x}'_1, \ldots, \mathbf{x}'_k\}$

\STATE Compute actionable features: $\mathcal{A} \leftarrow \{1, \ldots, p\} \setminus \mathcal{F}$
\STATE Identify binary actionable features: $\mathcal{B} \leftarrow \{j \in \mathcal{A} : |\text{unique}(x_j)| = 2\}$
\STATE Set $m \leftarrow |\mathcal{B}|$

\STATE 
\STATE \emph{// Exact Enumeration Branch}
\IF{$m \leq m_{\max}$ \AND $m > 0$ \AND $2^m - 1 \leq 2^{20}$}
    \STATE Initialize: $\mathcal{C}_{\text{enum}} \leftarrow \emptyset$
    \FOR{each non-empty subset $S \subseteq \mathcal{B}$}
        \STATE Construct $\mathbf{x}'_S$ by toggling features in $S$ according to Equation~(3)
        \STATE Compute predicted probability: $p_S \leftarrow f_{y^*}(\mathbf{x}'_S)$
        \IF{$p_S \in [p_{\min}, p_{\max}]$}
            \STATE Compute score: $s_S \leftarrow \alpha \cdot |S| - \beta \cdot |p_S - f_{y^*}(\mathbf{x}_0)|$
            \STATE Add to candidate set: $\mathcal{C}_{\text{enum}} \leftarrow \mathcal{C}_{\text{enum}} \cup \{(\mathbf{x}'_S, s_S)\}$
        \ENDIF
    \ENDFOR
    \IF{$|\mathcal{C}_{\text{enum}}| > 0$}
        \STATE Sort $\mathcal{C}_{\text{enum}}$ by score $s_S$ in ascending order
        \RETURN Top $k$ counterfactuals from $\mathcal{C}_{\text{enum}}$
    \ENDIF
\ENDIF

\STATE \emph{// Optimization Fallback: Stage 1 (NICE)}
\STATE Restrict training data: $\mathcal{D}_{\text{restricted}} \leftarrow \{\mathbf{x} \in \mathcal{D} : x_j = x_{0,j} \text{ for all } j \in \mathcal{F}\}$

\IF{$|\mathcal{D}_{\text{restricted}}| > 0$}
    \STATE Run NICE to find $k$ nearest neighbors in $\mathcal{D}_{\text{restricted}}$ using Gower distance
    \STATE Filter candidates where $f_{y^*}(\mathbf{x}') \in [p_{\min}, p_{\max}]$
    \STATE Validate that only features in $\mathcal{A}$ are modified
    \IF{valid counterfactuals found}
        \RETURN Top $k$ counterfactuals ranked by proximity and sparsity
    \ENDIF
\ENDIF

\STATE \emph{// Optimization Fallback: Stage 2 (MOC)}
\STATE Initialize MOC with population size $\mu = 40$ and $g = 60$ generations
\STATE Set fixed features $\mathcal{F}$ to remain unchanged during evolution
\STATE Evolve population optimizing: proximity, sparsity, and $f_{y^*}(\cdot)$
\STATE Filter final population: retain only $\{\mathbf{x}' : f_{y^*}(\mathbf{x}') \in [p_{\min}, p_{\max}]\}$
\STATE Validate actionability constraints
\RETURN Top $k$ counterfactuals from Pareto front
\end{algorithmic}
\end{algorithm}

\section{Results}

In the MGB P2RC, a total of 85{,}376 individuals were identified with 98{,}069 documented episodes of SARS-CoV-2 infection. Among these, 391 patients developed post-acute sequelae of COVID-19 presenting as heart failure (HF-PASC) within twelve months of infection, whereas 2{,}353 patients with prior HF did not meet PASC criteria (Table~\ref{tab:baseline}). Additional information regarding the curation and validation of this cohort is available in Azhir \textit{et~al.} (2024). 

\begin{table}[htbp]
\centering
\caption{Summary statistics of the study population}
\label{tab:baseline}
\begin{tabularx}{\linewidth}{>{\centering\arraybackslash}X
                                >{\centering\arraybackslash}c
                                >{\centering\arraybackslash}c}
\toprule
& \textbf{HF (N = 391)} & \textbf{Overall (N = 2744)} \\
\midrule
\textbf{Age} Mean (SD)   & 69.8 (14.7)  & 70.1 (13.9) \\
\textbf{Female}, n (\%)  & 183 (46.8\%) & 1273 (46.4\%) \\
\textbf{ECI} Mean (SD)   & 6.16 (2.87)  & 6.19 (2.98) \\
\bottomrule
\end{tabularx}
\end{table}

The overall cohort had a mean age of 70.1 years. Approximately 46.4\% of participants were female. The mean ECI was 6.19 across all participants and 6.16 among those who developed HF-PASC, reflecting a similar comorbidity burden between groups.

\begin{figure}[H]
    \centering
    \includegraphics[width=\linewidth]{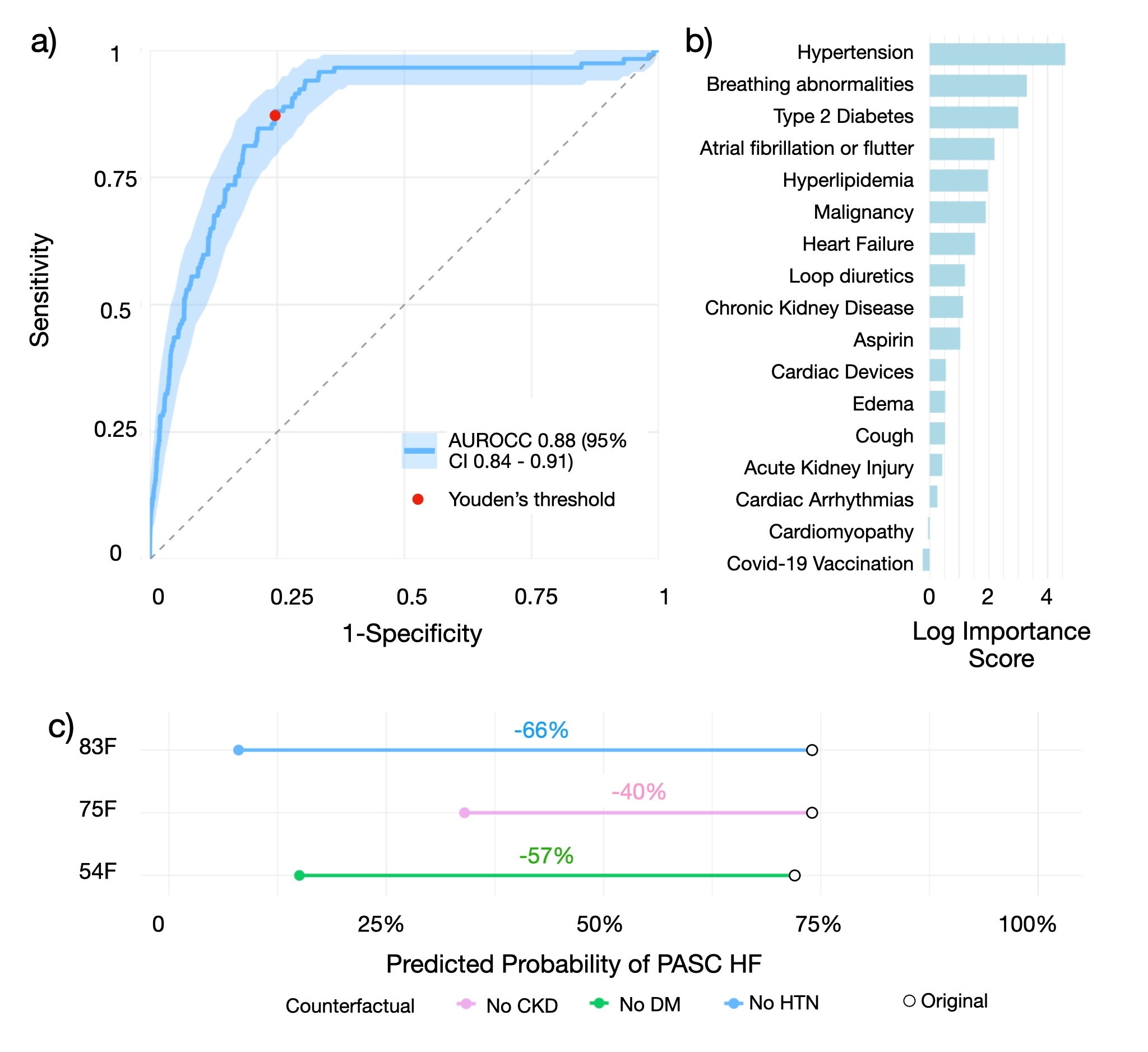}
    \captionsetup{justification=raggedright,singlelinecheck=false}
    \caption{
    (a) ROC for the test set with Youden’s threshold indicated; 
    AUROC = 0.88 (95\% CI: 0.84–0.91). 
    (b) Top model predictors ranked by relative importance. 
    (c) Patient-level counterfactuals illustrating PASC-HF risk modification following COVID-19 infection.
    Case 1 (blue): 83-year-old female with poorly controlled hypertension (HTN) and heart failure with preserved ejection fraction (HFpEF); 
    counterfactual scenario “no HTN” reduces predicted risk from 74\% to 8\%. 
    Case 2 (pink): 79-year-old female with HTN, HFpEF, coronary artery disease (CAD), and chronic kidney disease (CKD; Cr 1.3–1.8); 
    counterfactual “no CKD” decreases risk from 74\% to 34\%. 
    Case 3 (green): 54-year-old male with HTN and poorly controlled diabetes (A1c 9.3\%), complicated by retinopathy, nephropathy, neuropathy, and diabetic ketoacidosis (DKA); 
    counterfactual “no diabetes” lowers risk from 72\% to 15\%.
    Counterfactual analyses assume all other clinical features remain constant. 
    }
    \label{fig:roc_counterfactuals}
\end{figure}

The predictive model exhibited robust discriminative ability, achieving an area under the receiver operating characteristic curve (AUROC) of \( 0.88 \) (95\% CI: 0.84–0.91) (Figure~\ref{fig:roc_counterfactuals}). At the optimal operating threshold determined by Youden’s index (\( p^{*} = 0.41 \)), the model attained a sensitivity of \( 0.88 \) and a specificity of \( 0.76 \), indicating balanced performance across true-positive and true-negative classifications. 

Counterfactual inference analyses were subsequently performed to quantify the impact of modifiable clinical variables on predicted risk. 
Across representative test instances, transitioning specific comorbid conditions such as \textit{hypertension}, \textit{chronic kidney disease (CKD)}, or \textit{diabetes mellitus} from the present to the absent state resulted in marked reductions in estimated PASC-related heart failure risk (e.g., from approximately 72\% to 15\%). 
These counterfactual evaluations demonstrate the capacity of the proposed inference framework to reveal actionable, patient-level determinants of post-COVID outcomes within a rigorous probabilistic modeling context.

\section{Discussion}

This work demonstrates how counterfactual inference can serve as a unified framework for individualized prediction and causal reasoning in complex clinical data. By integrating regularized predictive modeling with both enumeration-based and optimization-based counterfactual search, our approach moves beyond population-level risk modeling to simulate patient-specific, hypothetical interventions. Applied to patients with pre-existing heart failure following COVID-19 infection, the model revealed interpretable relationships between comorbidities, laboratory findings, and PASC outcomes. Counterfactual analyses quantified how modifications in clinical features, such as reducing comorbidity burden or altering treatment patterns, could change predicted risks of PASC-related HF admission.

Our model accurately stratifies the risk of PASC-HF admission among HF patients and identifies actionable clinical targets that can inform individualized prevention strategies and mitigate the long-term burden of PASC-HF. These findings highlight the potential of counterfactual reasoning to bridge predictive accuracy and causal interpretability, enabling transparent, personalized inference that supports clinical decision-making and hypothesis generation.

In contrast to single-method implementations, our procedure is organized as a unified pipeline with a common objective and constraint set. When the actionable set is small, the method performs exhaustive enumeration, thereby exploring all feasible counterfactuals and returning solutions that are optimization-guaranteedunder the specified scoring rule. When the combinatorial space becomes intractable, the pipeline transitions to NICE and MOC to explore instance-grounded and multi-objective trade-offs while preserving consistency with the same objective and constraints. This structure provides deterministic solutions in tractable regimes and a principled approximation strategy otherwise, facilitating reproducible comparison across regimes within a single framework. A practical limitation of this design is the finite feature budget for which exhaustive enumeration is tractable. Beyond that threshold, the method necessarily defaults to NICE/MOC, trading guaranteed optimality for scalable approximation within the same framework.

This study uses TLDR to improve temporal windowing of EHR events, yielding cleaner, time-resolved features for diagnoses, medications, and laboratory findings, including scenarios with multiple treatments and multiple outcomes. TLDR sharpens the inputs, while counterfactual inference provides the core value by simulating patient-specific “what-if” interventions at a concrete clinical episode. This enables clinicians to see how targeted changes in a treatment plan could alter predicted risk, delivering actionable, individualized insight that standard atemporal models cannot.

A limitation of the current approach is that interpretability is strongest for binary features (presence/absence). Future work should extend the framework to continuous variables in clinically meaningful ways, for example, by incorporating predefined lab-value ranges, actionable thresholds, and dose/intensity levels, to provide more nuanced counterfactuals for numeric features. Strengthening this aspect would preserve the clarity of the current results while broadening their applicability to common continuous measurements used in routine care.

The counterfactual inference framework presented here has potential applications beyond individual risk prediction, particularly in supporting in-silico clinical trials and drug repurposing efforts. By modeling "what-if" scenarios where patients receive alternative medications or treatment regimens, the approach could simulate patient responses to different therapeutic interventions without requiring actual patient exposure. For drug repurposing, the framework could evaluate how existing medications might alter disease trajectories in patient subgroups by generating counterfactuals that modify medication features while preserving patient demographics and comorbidity patterns. Such simulations could help prioritize candidate drugs for formal clinical testing, potentially reducing the cost and timeline of drug development pipelines. However, several limitations must be acknowledged. For example, the framework assumes that feature modifications represent achievable clinical interventions, which may not hold for all therapeutic scenarios. Despite constraints, counterfactual modeling could serve as a valuable screening tool to identify promising therapeutic hypotheses for subsequent rigorous clinical evaluation, particularly when integrated with mechanistic understanding of disease pathways and drug action.

\section{Data Availability}  
Due to patient privacy regulations, the dataset is not publicly available. 

\section{Ethics approval}
The use of these data was authorized under the Mass General Brigham Institutional Review Board (IRB) protocol \#2020P001063, which granted a waiver of consent for this data-only retrospective study. MGB is an integrated healthcare delivery network serving approximately 1.5 million patients annually.

\section{Declaration of Interests}
The authors declare no competing interests.

\section{Acknowledgments}
This study has been supported by grants from the National Institute of Allergy and Infectious Diseases (NIAID) R01AI165535. 


\end{document}